\pdfoutput=1

\documentclass[11pt]{article}

\usepackage[preprint]{acl}

\usepackage{times}
\usepackage{latexsym}
\usepackage{float}

\usepackage{xcolor}
\usepackage{array}
\usepackage{booktabs}
\newcommand{\greenup}[1]{\textcolor{green}{#1\%$\uparrow$}}
\newcommand{\reddown}[1]{\textcolor{red}{#1\%$\downarrow$}}

\usepackage[T1]{fontenc}

\usepackage[utf8]{inputenc}

\usepackage{microtype}

\usepackage{inconsolata}

\usepackage{graphicx}

\title{Beneath the Surface of Consistency: Exploring Cross-lingual Knowledge Representation Sharing in LLMs}

\usepackage{amsfonts}
\usepackage{color}

\author{
 \textbf{Maxim Ifergan \textsuperscript{1}},
 \textbf{Leshem Choshen \textsuperscript{1}},
 \textbf{Roee Aharoni \textsuperscript{2}},
 \textbf{Idan Szpektor\textsuperscript{2}},
 \textbf{Omri Abend \textsuperscript{1}},
\\
 \textsuperscript{1} The Hebrew University of Jerusalem,
 \textsuperscript{2} Google Research,
\\
\texttt{\{{first name\}.\{last name\}@mail.huji.ac.il}}
 \\
 \texttt{\{roeeaharoni, szpektor\}@google.com}
}

\begin{document}
\maketitle

\begin{abstract}

The veracity of a factoid is largely independent of the language it is written in. However, language models are inconsistent in their ability to answer the same factual question across languages. This raises questions about how LLMs represent a given fact across languages.
We explore multilingual factual knowledge through two aspects: the model's ability to answer a query consistently across languages, and the ability to ''store'' answers in a shared representation for several languages.
We propose a methodology to measure the extent of representation sharing across languages by repurposing knowledge editing methods. We examine LLMs with various multilingual configurations using a new multilingual dataset. We reveal that high consistency does not necessarily imply shared representation, particularly for languages with different scripts. Moreover, we find that script similarity is a dominant factor in representation sharing.
Finally, we observe that if LLMs could fully share knowledge across languages, their accuracy in their best-performing language could benefit an increase of up to 150\% on average. These findings highlight the need for improved multilingual knowledge representation in LLMs and suggest a path for the development of more robust and consistent multilingual LLMs.

\end{abstract}

\section{Introduction}

\begin{figure}[t]
\centering
\includegraphics[scale=0.38]{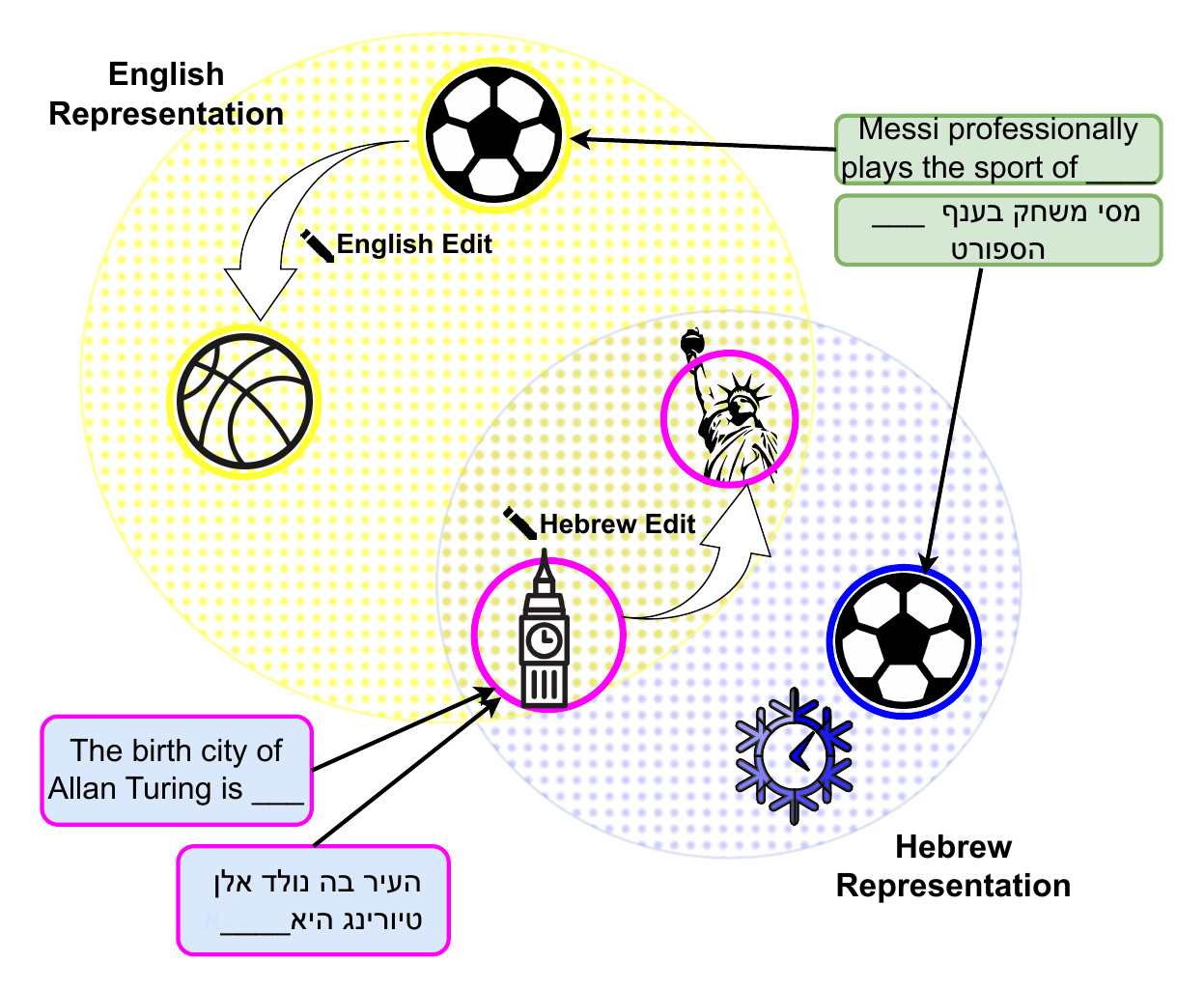}
\caption{Illustration of our method for distinguishing between cross-lingual consistency and representations sharing in a pairwise language setting. The sports (green) question demonstrates mere cross-lingual answer consistency, while the query about Allan Turing's birthplace (blue) exemplifies a shared underlying representation. Edits to the shared representation propagate across both languages, unlike the consistent-only fact. This method exposes the crucial difference between surface-level answer consistency and genuine cross-lingual knowledge sharing.}
\label{fig:pairwiseilustraion}
\end{figure}

Pretrained large language models (LLMs) have demonstrated a remarkable capacity to encode and retrieve factual knowledge \citep{Petroni2019LanguageMA,chang2024large} across diverse languages \citep{kassner2021multilingual, jiang2020x}. However, substantial variation in model performance across languages with a strong bias toward high-resource languages \citep{kassner2021multilingual, jiang2020x, fierro2022factual, jiang2022xlm, qi2023cross} highlights the issue of cross-lingual knowledge (in-)consistency.

This inconsistency raises questions about how LLMs represent factual knowledge in different languages. On one end of the range of possibilities, models may store a set of distinct knowledge copies for each language. On the other end, models may store a single, shared representation of the factual knowledge and ``decode'' it into surface forms in different languages. Thus, a shared representation manifests in consistency across languages, but consistent behavior can also occur without it. 

While consistency can be readily measured through agreement on identical queries across languages, measuring the extent to which knowledge representation is shared across languages requires more than just evaluating black box outputs. To quantify representation sharing, we propose editing factual knowledge in one language and examining the effects on other languages. For this purpose, we employ three knowledge editing techniques: two locate-and-edit methods, ROME \citep{Meng2022LocatingAE} and MEMIT \citep{Meng2022MassEditingMI}, and a finetuning-based method \cite{gangadhar2024model}. These methods are designed to surgically modify the relevant components of the model responsible for storing factual knowledge and only them, 
as illustrated in Fig~\ref{fig:pairwiseilustraion}.

As a test bed for our experiments, we compiled CLIKE, a multilingual ``fill-in-the-blank'' factual knowledge probing dataset with 35k samples, designed for evaluation and editing across paraphrases in 13 languages with 7 scripts. We experiment with diverse 7B-parameter LLMs supporting different sets of languages in different setups: monolingual, bilingual, multilingual, and language-extended models. Our analysis reveals a significant disparity in factual knowledge retrieval across languages. We assessed that on average these models answer correctly in at least one language 150$\%$ more facts than their best-performing language and triple the number of facts averaged across all 13 languages.

We assess for the first time the extent of cross-lingual knowledge representation sharing. We find that languages within the same script family exhibit the highest degree of representation sharing. This trend is consistent across all the models we studied, regardless of their level of multilingualism. Moreover, we find that high agreement in answers across certain language pairs does not entail shared internal representations, especially with language pairs that do not share the same script. This mismatch is particularly evident when comparing lower-resource languages to any language with a different script, which shows high consistency but limited representation sharing.

We expect that shedding light on these mechanisms will support the development of better multilingual models, with more efficient representation of factual knowledge. This will in turn lead to a more balanced knowledge across different languages, ultimately enhancing LLM performance across languages.


\section{Methodology}

In our analysis, we would like to measure two main aspects:

\begin{enumerate}
    \item \textbf{Cross-lingual Knowledge Consistency (CKC).} The extent to which a model shows consistency in answering factual questions when asked in different languages.
    \item \textbf{Cross-lingual Knowledge Representation Sharing (CKR).} The degree to which the model uses a common inner representation for the same fact across different languages.
\end{enumerate}

\subsection{Measuring CKC}

For simplicity, we say an LLM knows a fact in a specific language, modeled as a question and answer pair, if it can correctly answer it through a query written in this language. We start by defining a model's Knowledge Base (KB) for a specific language, as a set of facts an LLM 'knows' in that language.  Formally, for a given LLM $M$ and a dataset $D=\left\{f_{i}\right\}_{i\in [N]}$ of facts. Where $f_i^l:= (q_i^l, a_i^l)$ is a question-answer pair written in the language $l \in L$:
\[KB_{l}:=\{f_{i}\in D \mid\,M\left(q_{i}^{l}\right)=a_{i}^{l} \}.\]
To capture the pairwise relationship of knowing a fact in language $l_{1}$ to know it in $l_{2}$, we define $C_{(l_{1},l_{2})}$ as the conditional probability
\[P(f_{i}\in KB_{l_{2}}\mid f_{i}\in KB_{l_{1}})=\frac{\left|KB_{l_{1}}\cap KB_{l_{2}}\right|}{\left|KB_{l_{1}}\right|}. \]
We continue by defining the Number of Consistent Languages (NCL) a fact $f$ known in as:
\[NCL(f)=\left|\{l\in L:\,f\in KB_{l}\}\right|\]
With this aggregation, we can then compute the overall CKC of a model, as the average number of languages in which the LLM knows a fact:
\[\mathbb{E}[NCL]= \frac{1}{N} \sum_{f\in D} NCL(f)\]

\subsection{Measuring CKR}

Measuring the extent of shared knowledge representation across languages in LLMs cannot be done by merely evaluating model outputs. The same correct answer to a factual query across multiple languages could be generated from distinct, language-specific representations within the model, rather than a unified, language-agnostic abstraction. This requires a more sophisticated approach.

To measure this, we use an editing method $E$ that modifies the model's parameters to provide a wrong answer for a query in a given language. We then examine the impact of such a change on the same fact query in other languages. Let $M_i^l$ denote the updated model applying $E$ to the fact $f_i^l$ in the language $l$ to the target answer $t_i^l$. The model's KB for a specific language $l'$ after the modification in language $l$ is defined as the collection of facts for which the incorrect target answer, edited in language $l$, also propagates to language $l'$, which can be formally expressed as:
\[KB_{l'}^{l}:=\{f_i \in KB_{l_{1}}: M_i^{l_{1}}\left(q_i^{l_{2}}\right)=t^{l_{2}}_i\}.\]
We can then estimate the amount of pairwise CKR between a language $l_1$ to $l_2$ by defining $SR_{(l_{1},l_{2})}$ as the conditional probability
\[P(f\in KB_{l_{2}}^{l_{1}}\mid f\in KB_{l_{1}}^{l_{1}}) =\frac{\left|KB_{l_{1}}^{l_{2}}\cap KB_{l_{1}}^{l_{1}}\right|}{\left|KB_{l_{1}}^{l_{1}}\right|}.\]
We further define the Number of Transferred Languages (NTL) for a given fact edited in the language $l$ as
\[NTL(f^{l})=\left|\{l': f\in KB_{l'}^{l}\}\right|\]
With this aggregation, we can then compute overall CKR, as the average number of languages in which the LLM represents a fact as
\[\mathbb{E}[NTL]= \frac{1}{|L|} \sum_{l \in L} \left(\frac{1}{|KB_l|} \sum_{f\in KB_l} NTL(f^{l})\right).\]

\section{Experimental Setup}

\subsection{Data}

\textbf{Dataset.} 
We present CLIKE (Cross-LIngual Knowledge Editing), a dataset for evaluating and editing factual knowledge of pretrained LMs across languages and paraphrased expressions. CLIKE contains approximately 35k facts spanning 13 languages: English (en), French (fr), Italian (it), Spanish (es), Russian (ru), Ukrainian (uk), Bulgarian (bg), Hindi (hi), Bengali (bn), Chinese (zh), Japanese (ja), Hebrew (he), and Arabic (ar). Each fact is modeled as a language-independent (subject, relation, object) triplet and each relation has 3 paraphrased natural language templates for every language. Each template forms a sentence that conveys a fact and ends with the object, which we omit and expect the model to fill. 

For example, the triplet $(Bach, Birth City, Leipzig)$ will be converted to 'Bach was born in the city of' and 'The birth city of Bach is', and 'The birthplace of Bach is the city of' expecting the pretrained LM to complete the prompt with 'Leipzig' correctly using its initial pretraining task without altering the model with a finetuning intervention.

\paragraph{Fact Collection.} 
Following a similar approach to \citet{Petroni2019LanguageMA, sciavolino-etal-2021-simple, kassner2021multilingual, Wei2024MLaKEMK}, fact triplets were collected from \href{https://query.wikidata.org/}{Wikidata Query Service}. We manually crafted and published 14 SPARQL relation queries. Each query extracts wikidata entries for subjects and objects satisfying the query relation with their labels in all available languages. 
We then filtered all triplets with labels containing less than 8 of the examined languages to balance the languages in the dataset. Appendix~\ref{sec:CLIKE Dataset Key statistics} includes the languages and relation distributions.

\paragraph{Dataset Construction.} We used ''Gemini Advanced'' and ''Claude Opus'' to generate the templates of each relation in all languages. For each relation, we generated 3 paraphrases adjusted to grammar rules such as the gender of the subject. The prompts for these templates were executed on the models' official websites (\href{https://gemini.google.com/app}{Gemini}, \href{https://claude.ai/new}{Claude}).
Subsequently, professional translators or native speakers refined the templates and sampled generated fill-in-the-blank queries across all languages, following instructions detailed in Appendix~\ref{sec:NativeSpeakerInstruction}. For the knowledge editing task, we generated false but plausible objects for each fact by randomly sampling from other facts within the same relation category. This approach provided consistent incorrect alternatives across all languages for each query.

\subsection{Models}

We examine a range of LLMs with 7B parameters and decoder-only architectures. We focus on base pretrained language models to capture the knowledge acquired during the pretraining process, prior to any finetuning. BLOOM-7B \citep{Scao2022BLOOMA1} serves as our \textbf{multilingual} model. Qwen-7B \cite{Bai2023QwenTR} represents a \textbf{bilingual} Chinese-English model with a low tokenization compression rate multilingual vocabulary. We include two \textbf{monolingual} English models: Llama-2-7B \citep{Touvron2023Llama2O} and Mistral-7B-v0.1 \citep{Jiang2023Mistral7}. Additionally, we examine two \textbf{language-extended models}, \href{https://huggingface.co/hfl/chinese-llama-2-7b}{Chinese-llama-2-7B}, and \href{https://huggingface.co/yam-peleg/Hebrew-Mistral-7B}{Hebrew-Mistral-7B}, based on Llama-2-7B and Mistral-7B-v0.1 with additional pretraining in English and their expanded language (EL) and an expand EL tokenizer vocabulary. These models represent a diverse set of multilingual configurations, enabling a extensive analysis of cross-lingual knowledge representation.

\subsection{Knowledge Editing Methods}

We employ three knowledge editing methods: Finetuning (FT) \cite{gangadhar2024model}, ROME \citep{Meng2022LocatingAE}, and MEMIT \citep{Meng2022MassEditingMI}.
The ROME and MEMIT editing methods leverage causal mediation analysis \citep{vig2020causal, vig2020investigating} to identify the LM layer that has causally contributed to factual knowledge recall, suggesting the middle MLP layers act as key-value associative memory. ROME then computes a closed-form rank-one update to the layer's weights, inserting a new fact while minimizing disruption to existing knowledge stored in the weights. Similarly, MEMIT identifies a range of MLP layers that jointly contribute to the model's factual associations. Then it iteratively updates the weights of each MLP layer, distributing the changes across the MLP layers.

Both ROME and MEMIT use interpretability techniques to precisely locate and surgically modify the relevant components of the model responsible for storing factual knowledge. This approach allows for direct control over the model's memorized information while preserving its overall capabilities, providing a framework for isolating changes in actual knowledge without altering other components.

Finetuning, our baseline approach, involves updating the weights of all middle layers in the model without the MLP restrictions imposed by ROME and MEMIT. For each fact to be edited, we finetuned the model on a single example consisting of the edition prompt paired with its new target answer. It incorporates new factual knowledge that resembles standard language model training practices.

We use the EasyEdit code library \cite{wang2023easyedit} to perform all language model knowledge edits. Default parameters are employed for all models except BLOOM. Since BLOOM lacks a pre-existing implementation, we optimized and published custom hyperparameters for the editing methods.

\subsection{Metrics and Evaluation}

We employed the Exact Match (EM) metric to evaluate all answers to queries across our experiments. To provide context for the pretrained LLM, we used 3-demonstrations fewshot concatenated facts for both evaluation and editing tasks, maintaining the same examples and order. All answer generation was performed using greedy decoding to ensure deterministic outputs.

Model performance and CKC in a given language were assessed as follows. The overall accuracy for a language was computed as the percentage of facts correctly answered in at least one paraphrased form. $C(l_1,l_2)$, was measured by computing the mean score in language $l_2$ across all paraphrases for facts known in language $l_1$. Similarly, within-language consistency, $C(l,l)$, was computed using the same approach, evaluating the model's consistency across paraphrases within a single language.

For the knowledge editing experiments, we randomly selected 500 known facts in each language to modify. We assessed the effectiveness of these edits using three standard metrics: {\it Reliability}, {\it Generalization}, and {\it Locality}. Reliability measures the accuracy of the model on the edited prompt itself. Generalization, denoted with $SR(l1, l2)$, evaluates the mean score across all paraphrases of the edited fact in all languages, including the language in which the edit was made (l1). This quantifies how well the edit transfers across both languages and paraphrases variations of the fact. For Locality testing, we randomly sampled known facts for each language and evaluated the model's mean accuracy to answer these unrelated queries correctly. This ensured that the edits did not negatively impact other knowledge in different languages.

\section{Results}

Before presenting our main findings, we first validate the methodology's performance. We found a strong correlation (0.87) between the results obtained from different knowledge editing methods. This consistency across various editing techniques suggests that our findings are robust and not reliant on a specific method. Given this consistency, we primarily present results using MEMIT, with other methods' results available in  Appendix~\ref{sec:pairwise_languages_relation_Appendix}. Additionally, all knowledge editing methods maintained high locality scores (averaging above 70\%), indicating that edits were specific and preserved broader model knowledge. Furthermore, all models exhibit some variation in performance across paraphrases even within the same language, aligning with findings from \citet{mizrahi2024state} and further justifying our approach of assessing knowledge using multiple paraphrases.

\begin{figure}[t]
\centering
\includegraphics[scale=0.37]{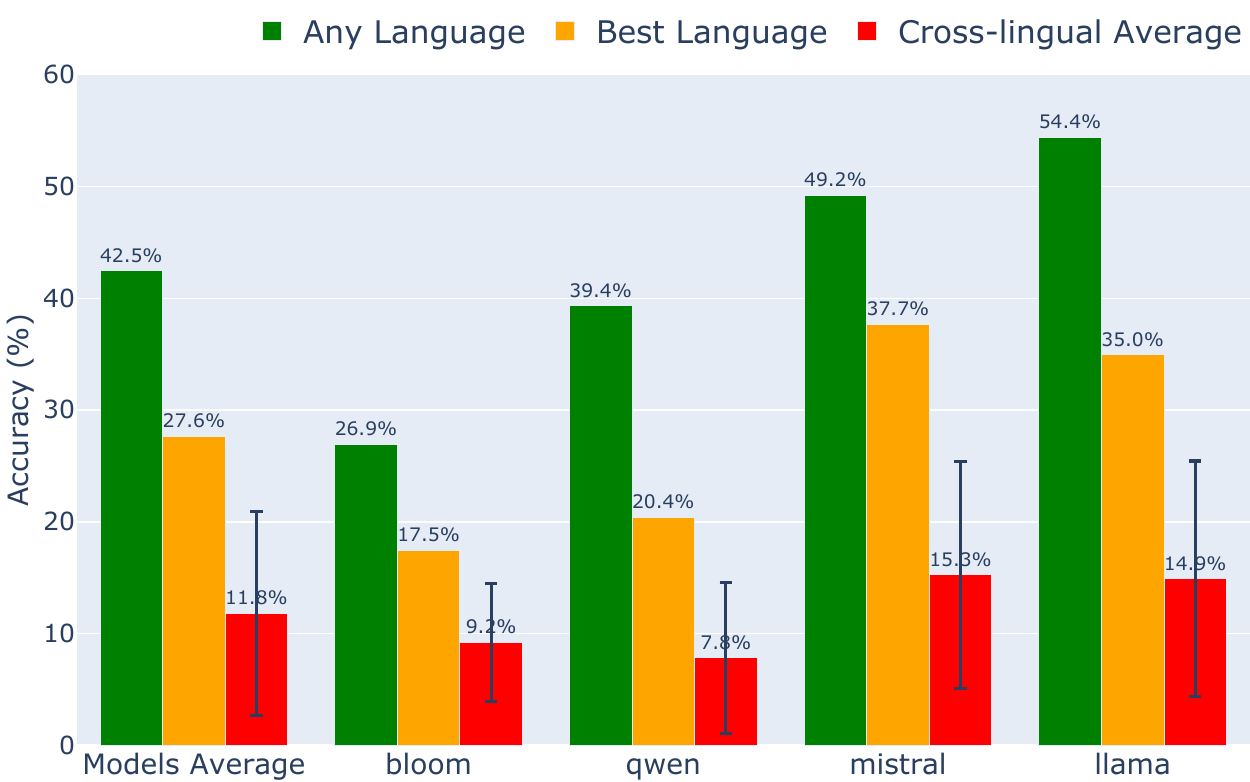}
\caption{Cross-lingual performance variability: the accuracy of factual knowledge retrieval across different languages for several LLMs supporting different language sets. 'Any Language' (green) -- facts known in at least one language, 'Best Language' (orange) -- accuracy in the best-performing language, and 'Cross-lingual Average' -- mean accuracy across all 13 languages in the CLIKE dataset, with error bars indicating standard deviation.}
\label{fig:Variability}
\end{figure}

\subsection{The Issue of Knowledge Variability}

Large language models (LLMs) exhibit significant variability in their factual knowledge retrieval across different languages, as illustrated in Fig.~\ref{fig:Variability}. Our analysis of four 7B-parameter LLMs reveals a striking disparity: while models demonstrate knowledge of 42.5\% of the facts on average in at least one language, their best-performing language achieves only 27.6\% accuracy, and their average performance across all 13 languages in the CLIKE dataset is merely 11.8\%.

If models could share knowledge across all languages, the best-performing language could potentially increase its accuracy by up to 53\%. Moreover, models could then potentially more than triple their current cross-lingual average accuracy. This observation motivates our subsequent investigations into CKR, as we seek to understand and potentially leverage these untapped reservoirs of knowledge.

\begin{figure*}[ht!]
  \includegraphics[width=0.66\linewidth]{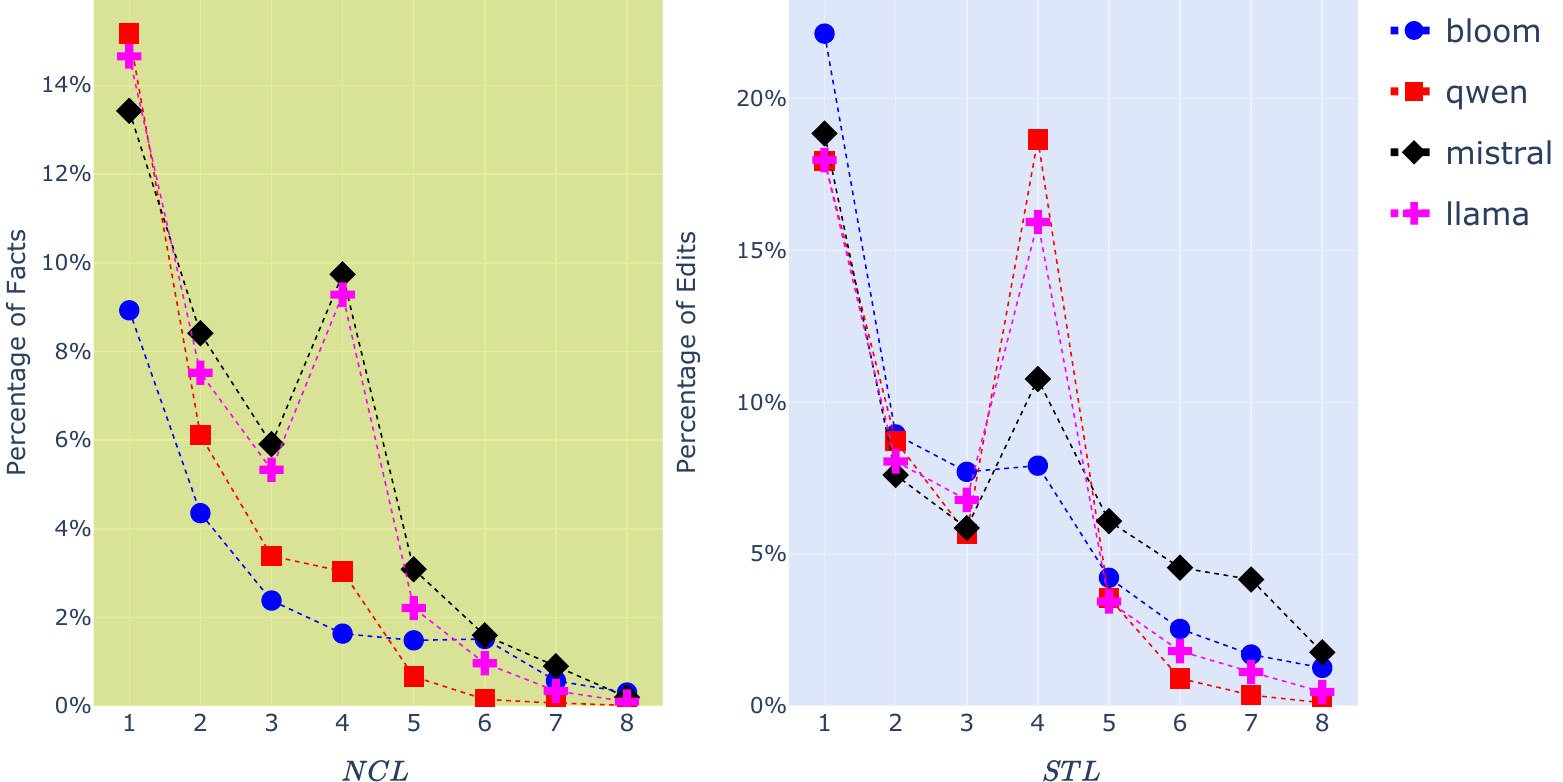} \hfill
  \includegraphics[width=0.33\linewidth]{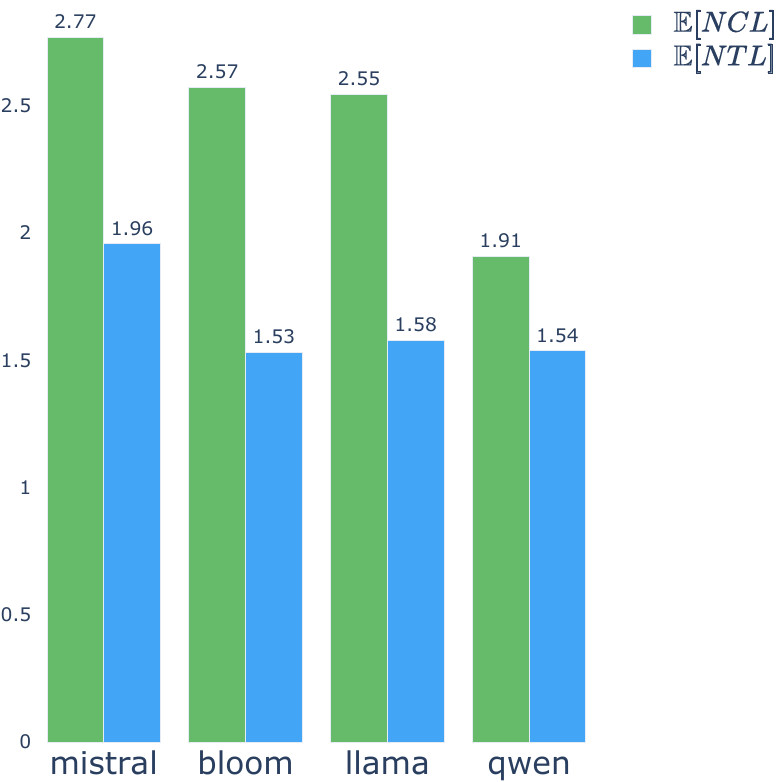}
  \caption {Distribution and Expectation of CKC and CKR. Left: Percentage of facts known (NCL) or represented (NTL) across multiple languages for different models. Right: Expected number of languages per fact (E[NCL]) and expected number of languages sharing representation per edited fact (E[NTL]) for each model, illustrating the relationship between knowledge CKC and CKR.
  }
\label{fig: NTL_ncl}
\end{figure*}

\begin{figure*}[ht!]
\centering
\includegraphics[scale=0.46]{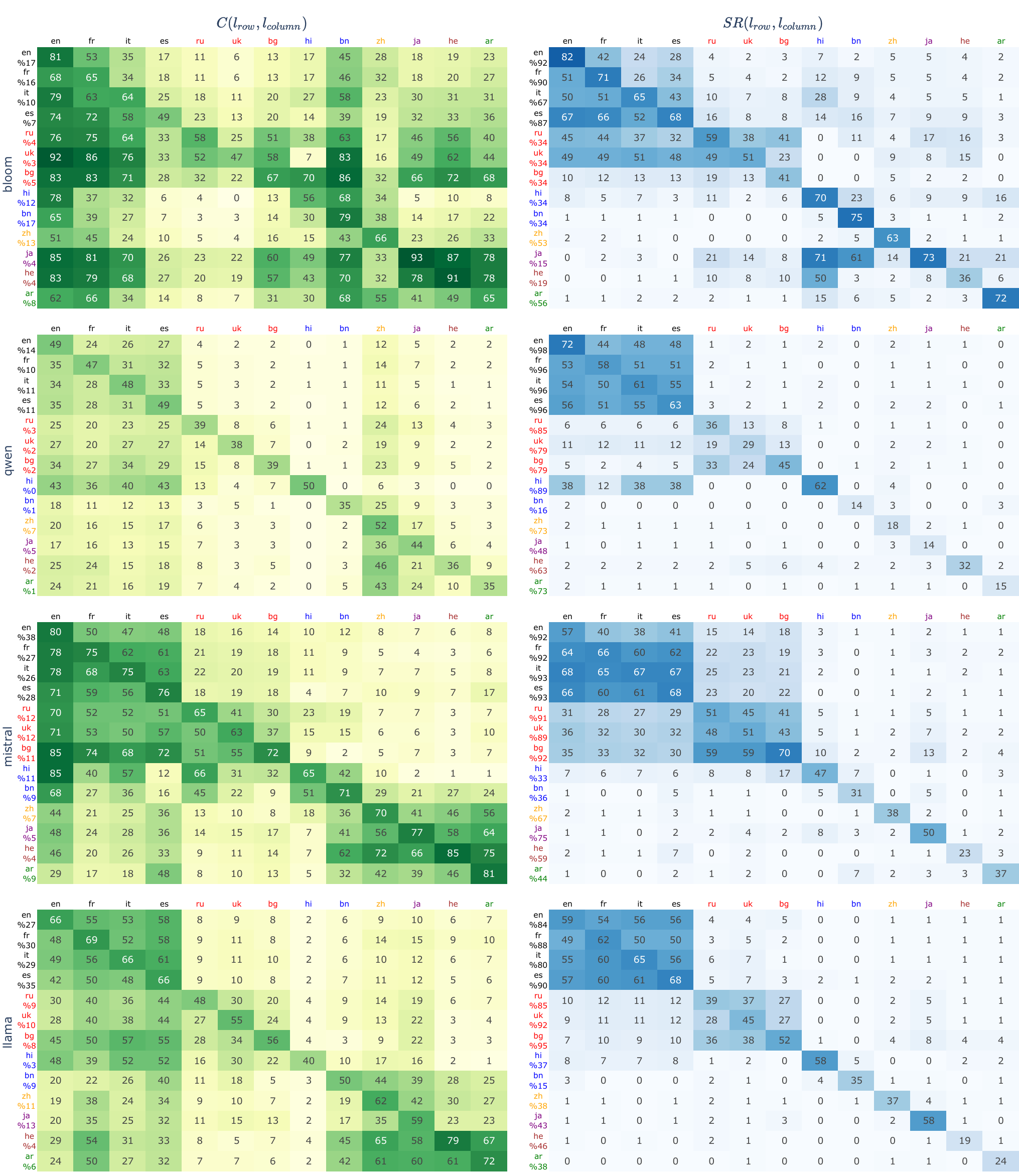}
\caption{
The pairwise relationships of factual knowledge across languages for four language models, with all scores reported using Exact Match. $C(l_1,l_2)$ shows the percentage of facts known in the row language which were also retrieved in the column language. $SR(l_1,l_2)$ indicates the percentage of successfully edited facts in the row language which generalized to the column language using MEMIT. Under each language abbreviation is the overall accuracy for initial knowledge retrieval and the edition-reliability score for $C$ and $SR$ measures respectively. Languages are color-coded by script family.}
\label{fig: pairwise_languages_relation}
\end{figure*}

\subsection{Consistency Does Not Imply Representation Sharing}

We decouple CKC and CKR between languages, examining both general measures across languages (Fig.~\ref{fig: NTL_ncl}) and pairwise language relationships addressing their specific identities (Fig.~\ref{fig: pairwise_languages_relation}). Our analysis reveals that high CKC does not necessarily imply high CKR, and in some cases, we observe inverse patterns.


At the general level, we observe for all models that $\mathbb{E}[NCL]$ is consistently higher than $\mathbb{E}[NTL]$, indicating that models tend to exhibit CKC across more languages than they share representations between. Interestingly, while $\mathbb{E}[NCL]$ values show considerable variation across models, $\mathbb{E}[NTL]$ values are more uniform. The persistent gap between $\mathbb{E}[NCL]$ and $\mathbb{E}[NTL]$ across all models highlights that consistent answers do not necessarily translate to shared internal representations. Moreover, we find that models with a lower proportion of facts known in only one language ($NCL$ = 1) tend to have a higher proportion of facts represented in only one language ($NTL$ = 1) as shown in Fig.~\ref{fig: NTL_ncl}.

At the pairwise language level, we find differences between CKC and CKR patterns. For instance, most models (except Qwen) exhibit a high degree of CKC among low-resource languages with different scripts (Chinese, Japanese, Hebrew, Arabic). However, when examining CKR, we find limited evidence of shared encoding between these languages. Conversely, we observe a higher degree of shared representation among Cyrillic languages compared to the shared representation between Cyrillic and Latin languages. This is despite the fact that CKC scores show an opposite trend, with higher CKC between Cyrillic and Latin languages than among Cyrillic languages themselves.

\subsection{The Key Role of the Language Script}

Our analysis provides quantifiable measures of CKR in LLMs. Following previous work \cite{qi2023cross, beniwal2024cross}, our study highlights the importance of the script of a language for multilingual knowledge. We observe that the pairwise $SR$ measure is relatively consistent across models, despite their varying language support.

We find that languages within the same script family exhibit the highest degree of CKR across all models. As shown in Fig.~\ref{fig: pairwise_languages_relation}, we observe a script-based grouping in both CKC and CKR likely highlighting a tokenization induced bias \citep{singh2019bert}. Notably, we observe strong CKR between languages with Latin scripts (English, French, Italian, Spanish) and between languages with Cyrillic scripts (Russian, Ukrainian, Bulgarian). For Devanagari script languages (Hindi, Bengali) we observe relatively high CKR for models that perform well on these languages (Bloom, Mistral).

While most CKR occurs among languages that use the same script, there is still some knowledge transfer between languages with different scripts. This cross-script transfer is particularly evident between Cyrillic and Latin script languages across various models. Additionally, in specific cases, such as with the BLOOM model, we observe a moderate degree of CKR between seemingly unrelated language pairs, e.g., 28\% from Italian to Hindi.

Notably, these relations between language scripts are sometimes asymmetrical. For example, knowledge in Cyrillic script languages implies a higher probability (approximately 40-60\%) of knowing the same facts in Latin script languages. However, the reverse relation is weaker, with only about 10-20\% probability of Cyrillic knowledge given Latin script knowledge. A similar asymmetrical relation appears across models suggesting a stronger transfer of knowledge from Cyrillic to Latin. We hypothesize that the dominance of Latin script languages, especially English, in the training data leads to more robust fact representations in Latin scripts, facilitating easier transfer from Cyrillic to Latin than vice versa.

\subsection{Impact of Model Design Languages}

How does a model's designed language support affect its CKR and CKC patterns? Although the patterns of CKR are relatively similar across models supporting different language sets, our analysis reveals some nuanced differences. 

The multilingual BLOOM model demonstrates the highest pairwise average of language pairwise CKC (36\%) and CKR (8.4\%) across different script pairs. As shown in Fig.~\ref{fig: pairwise_languages_relation}, BLOOM exhibits notable transfer between seemingly unrelated language pairs. These cross-script patterns validate BLOOM's design as a multilingual model, emphasizing its cross-lingual relationships rather than its overall low accuracy performance.

We find that the bilingual English-Chinese Qwen model is showing relatively high overall accuracy in Chinese. However, this Chinese knowledge remains largely distinct from English both in terms of language pairwise CKC and CKR Fig.~\ref{fig: pairwise_languages_relation}. This pattern validates Qwen's design as a bilingual model, emphasizing its language-specific capabilities rather than cross-script knowledge sharing. Surprisingly, when examining the global shared representation, Qwen exhibits a higher number of 4-lingual representations sharing (NTL = 4) from uniquely represented facts. Although Qwen lacks cross-script knowledge sharing, it developed some degree of multilingual representation, particularly within script families.

Monolingual English models (Mistral, LLaMA) exhibit a unique pattern. We discover an anomalous peak in the results for facts known and represented in exactly four languages (Fig.~\ref{fig: NTL_ncl}), corresponding primarily to the four Latin script languages in our dataset. This highlights the strong association between script similarity and knowledge sharing even in ostensibly monolingual models. Surprisingly, Mistral demonstrates the highest $\mathbb{E}[NCL]$, $\mathbb{E}[NTL]$ and average of language pairwise CKC (54.7\%) and CKR (37.6\%) within script families, despite being designed as a monolingual English model. This result highlights how Mistral's strong English foundation naturally extends to other Latin script languages, underscoring the impact of script similarity on cross-lingual knowledge representation even in monolingual models.


\begin{table}[ht!]
\scalebox{0.94}{
\small
\centering
\tabcolsep=0.06cm
\begin{tabular}{|c|c|c|c|c|c|}
\hline
 & Model & EL Acc/Rel & En Acc/Rel & EL $\rightarrow$ en & en $\rightarrow$ EL \\
\hline \hline
$C$ & zh EL & 10 (\greenup{142}) & 5 (\reddown{29}) & 12 (\reddown{80}) & 22 (\greenup{440}) \\
\hline
$C$ &  he EL & 18 (\greenup{600}) & 10 (\reddown{32}) & 13 (\reddown{37}) & 18 (\greenup{900}) \\
\hline\hline
$SR$ & zh EL & 96 (\greenup{252}) & 81 (\reddown{96}) & 2 (\greenup{200}) & 4 (\greenup{400}) \\
\hline
$SR$ & he EL & 90 (\greenup{152}) & 83 (\reddown{90}) & 10 (\greenup{500}) & 6 (\greenup{600}) \\
\hline
\end{tabular}
}
\caption{CKC and CKR in extended LLMs compared to their base models. EL: Extended Language, Acc: Accuracy, and Rel: Reliability.}
\label{tab:EL}
\end{table}

\subsection{Language Extended LMs}

How does additional pretraining on both English and an extended language (EL) impact cross-lingual CKC and shared representation in initially monolingual models? Analysis of chinese-llama-2-7b and he-mistral-7b reveals a similar trade-off: while gaining substantial knowledge in EL, models sacrifice much of their original English expertise. These extensions reshape cross-lingual knowledge distribution but fall short of fully bridging the gap between disparate writing systems.

As shown in Table \ref{tab:EL}, both models exhibit increased accuracy in the extended language (EL) coupled with decreased English accuracy. CKC measures paint a nuanced picture: models acquire extensive new knowledge in EL, largely unknown in English, yet this new EL knowledge covers more of English knowledge. Shared representation metrics underscore this asymmetry. Despite increased bidirectional knowledge transfer between English and EL, transfer remains stubbornly low. This suggests that even with targeted pretraining, models struggle to forge robust representations sharing across linguistically distant languages.

For analysis of how different relation types affect CKR, see Appendix~\ref{sec:ckr_features}.

\section{Related Work}

\paragraph{Cross-lingual Knowledge Consistency.}


While monolingual knowledge consistency has been studied often in LMs \citep{Elazar2021MeasuringAI, mizrahi2024state}, limited work has been done on cross-lingual knowledge consistency. \citet{qi-etal-2023-cross} proposed a cross-lingual consistency metric named RankC to measure similarity across multiple candidate answers, whether correct or incorrect. Our focus on correct answers allows a simpler assessment without being limited to pairwise language comparisons.

\paragraph{Cross-lingual Knowledge Representation Sharing.} Previous studies explored this angle through different approaches. Some works studied parameter sharing across languages by analyzing neuron activation/deactivation when evaluating knowledge in different languages \cite{libovicky2020language, zhao2024large, chen2024journey, tang2024language, kojima2024multilingual}. Enhancing language-independent neurons resulted in better multilingual abilities in a specific language without compromising others. Other works investigated the knowledge related to the training data and identified the language source of the acquired data \citep{choenni2023languages, zhao2024tracing}, providing evidence that knowledge from training data in one language can benefit the model in other languages. Another line of work analyzed how inputs in different languages affect the activation patterns, showing that semantically equivalent content in different languages tends to produce similar activation patterns \citep{singh2019bert, libovicky2020language, chang2022geometry}. 

These works pointed to a connection between knowledge in different languages. However, they do not yield an assessment of the amount of shared knowledge. While passive analysis can take as far as measuring the similarity between languages, active modification tools can also suggest a clear causal relation between the knowledge representation in different languages.

\paragraph{Multilingual Knowledge Editing.}
Previous work on multilingual knowledge editing \cite{Si2024MPNLM, Xu2022LanguageAC, Wei2024MLaKEMK, Wang2023CrossLingualKE} primarily focused on comparing and improving editing methods' performance in multilingual settings. Our approach is different. We use these editing tools as analytical tools to understand representation sharing across languages and across models with different multilingual configurations. 

\section{Conclusion}

This work investigated the relationship between cross-lingual knowledge consistency and representation sharing in LLMs. 
Our findings reveal that high consistency across languages does not necessarily imply shared internal representations, particularly for languages with different scripts. 
We introduced a novel methodology and dataset for quantifying these phenomena, providing a more nuanced understanding of how LLMs represent and retrieve factual knowledge. 
The significant disparity we observed in factual knowledge retrieval across languages, coupled with the potential for substantial performance improvements if knowledge could be fully shared, underscores the importance of developing more effective multilingual knowledge representations. 
We expect that our insights will guide the development of more efficient and equitable multilingual models, ultimately enhancing their performance across all languages.

\section*{Limitations}
Our main limitation lies in the constraints imposed by our chosen editing methods and their focus on specific model components. By primarily targeting middle layers associated with factual knowledge storage, our analysis may have overlooked important cross-lingual interactions occurring elsewhere in the model architecture. Our reliance on specific editing techniques (ROME, MEMIT, and Finetuning) may not capture the full spectrum of knowledge representation and modification within the model. There might be multiple pathways to change the output in a specific language, potentially exhibiting different cross-lingual generalization patterns than those we observed.

Our analysis focused exclusively on decoder-only language models with 7B parameters, limiting the generalizability of our findings across different architectures and sizes. Similarly, while our CLIKE dataset covers a diverse range of languages and relations, it may not fully represent the breadth of factual knowledge or linguistic phenomena. These constraints in both model selection and dataset composition could influence the observed patterns of cross-lingual representation.

\bibliography{custom}

\appendix
\clearpage
\newpage
\section{Native Speaker Instruction}
\label{sec:NativeSpeakerInstruction}
\fbox{\begin{minipage}{\textwidth}
Dear <Native Speaker>, \\\\
We are reaching out to you for assistance in an important project that aims to improve the ability of Artificial Intelligence (AI) to understand and generate text in your native language. Your skills and knowledge as a native speaker are crucial to the success of this project. \\
Our research team has created a collection of fill-in-the-blank sentences and templates in multiple languages, including yours. These sentences will be used to evaluate the knowledge and understanding of AI language models. To ensure the accuracy and effectiveness of our collection, we need your help in verifying the grammatical correctness of the sentences and templates we have created. \\
Attached, you will find a list of approximately 60 simple sentences and sentence templates and templates that cover various relationships between subjects and objects in your native language. The task should take no more than 15-20 minutes to complete. Your task is to review each sentence and template and determine whether they are grammatically correct. If you find any grammatical errors, please provide a corrected version of the template. Additionally, if you wish, you may provide an optional explanation in English of what was wrong with the original template. \\\\
\underline{Example:} \\\\
Relation: \textit{Birth City}, Subject: \textit{Wolfgang Amadeus Mozart}, and Object: \textit{Salzburg} \\
Original template: \\
\textit{"[subj] birthplace the city [obj]"} ->  \textit{"Wolfgang Amadeus Mozart birthplace the city Salzburg"} \\
Fixed template: \\
\textit{"[subj]'s birthplace is the city of [obj]"} -> \textit{"Wolfgang Amadeus Mozart's birthplace is the city of Salzburg"} \\
Explanation: \\
The original template is missing the verb "is" and the preposition "of" to form a grammatically correct sentence. \\\\
When fixing the templates, please keep in mind the following guidelines:
\begin{enumerate}
    \item Be explicit about the relationship to avoid ambiguity. For example, given the information (Bach, Birth Year, 1685) and the template "[subj] born in [obj]", the AI might complete the prompt "Bach was born in" with the object "Leipzig" (his birth city) or "31 March" (his birth date) rather than the year "1685". Therefore, a good template should contain words that explicitly describe the relationship. The template "Bach was born in the year [obj]" will likely output "1685" since the word "year" appeared in the sentence.
    \item For each relationship, we have provided three different prompt paraphrases that are supposed to be different from one another.
    \item The subject must always appear before the object.
    \item The last word of the prompt template should be the object.
\end{enumerate}
Please note that the sentences do not necessarily need to sound natural or be well-written. Our primary focus is on ensuring that they are grammatically correct. However, if you have suggestions on how to make the sentences sound more natural without changing the core structure, feel free to include them in your feedback. \\
Your contribution to this project is valuable and will help us create a reliable collection of sentences for advancing AI's ability to understand and generate text in your language. If you have any questions or concerns about the project or your role in it, please don't hesitate to reach out to us. \\
Thank you for your participation in this project. We look forward to receiving your feedback. \\
Best regards,
\end{minipage}}

\newpage
\clearpage
\section{CLIKE Dataset Key statistics.}
\label{sec:CLIKE Dataset Key statistics}
\begin{figure}[ht]
\centering
\includegraphics[scale=0.44]{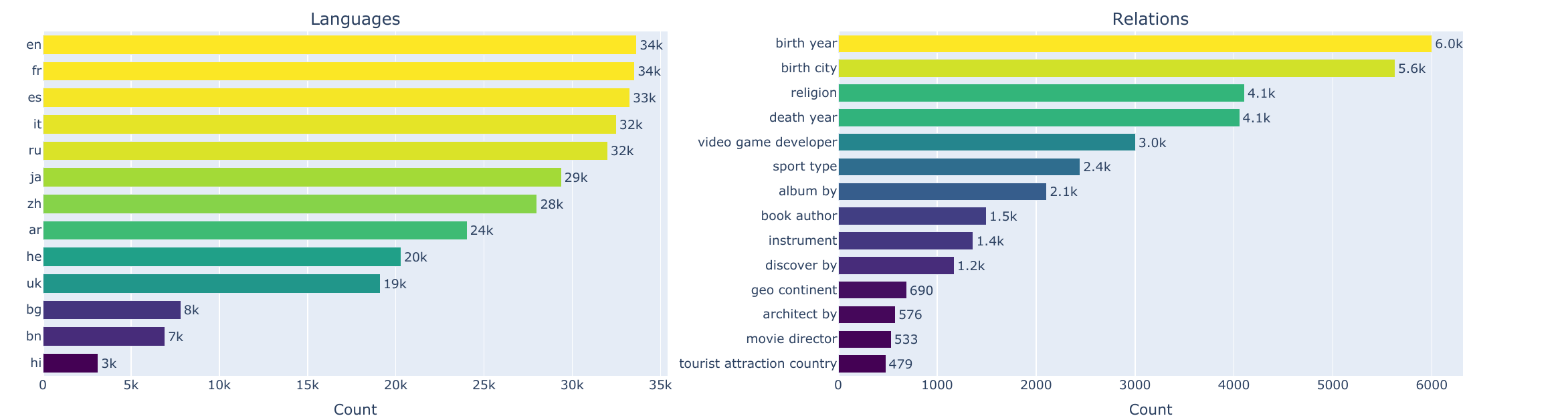}
\caption{Left: histogram of number of examples for each language, right: histogram of number of examples for each relation type.} 
\end{figure}

\begin{figure}[ht]
\centering
\includegraphics[scale=0.42]{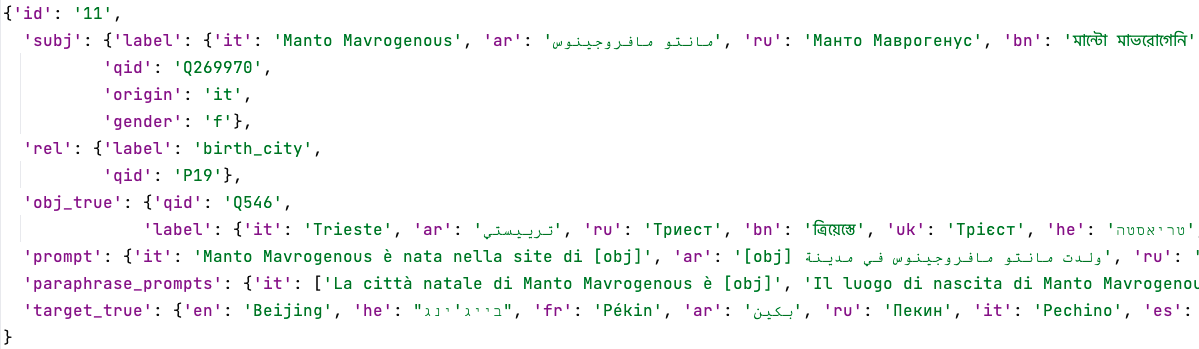}
\caption{A Dataset Sample Example} 
\end{figure}

\clearpage
\clearpage
\section{ROME and FT Methods Performance Comparison}
\label{sec:pairwise_languages_relation_Appendix}
\begin{figure}[ht!]
\centering
\includegraphics[scale=0.48]{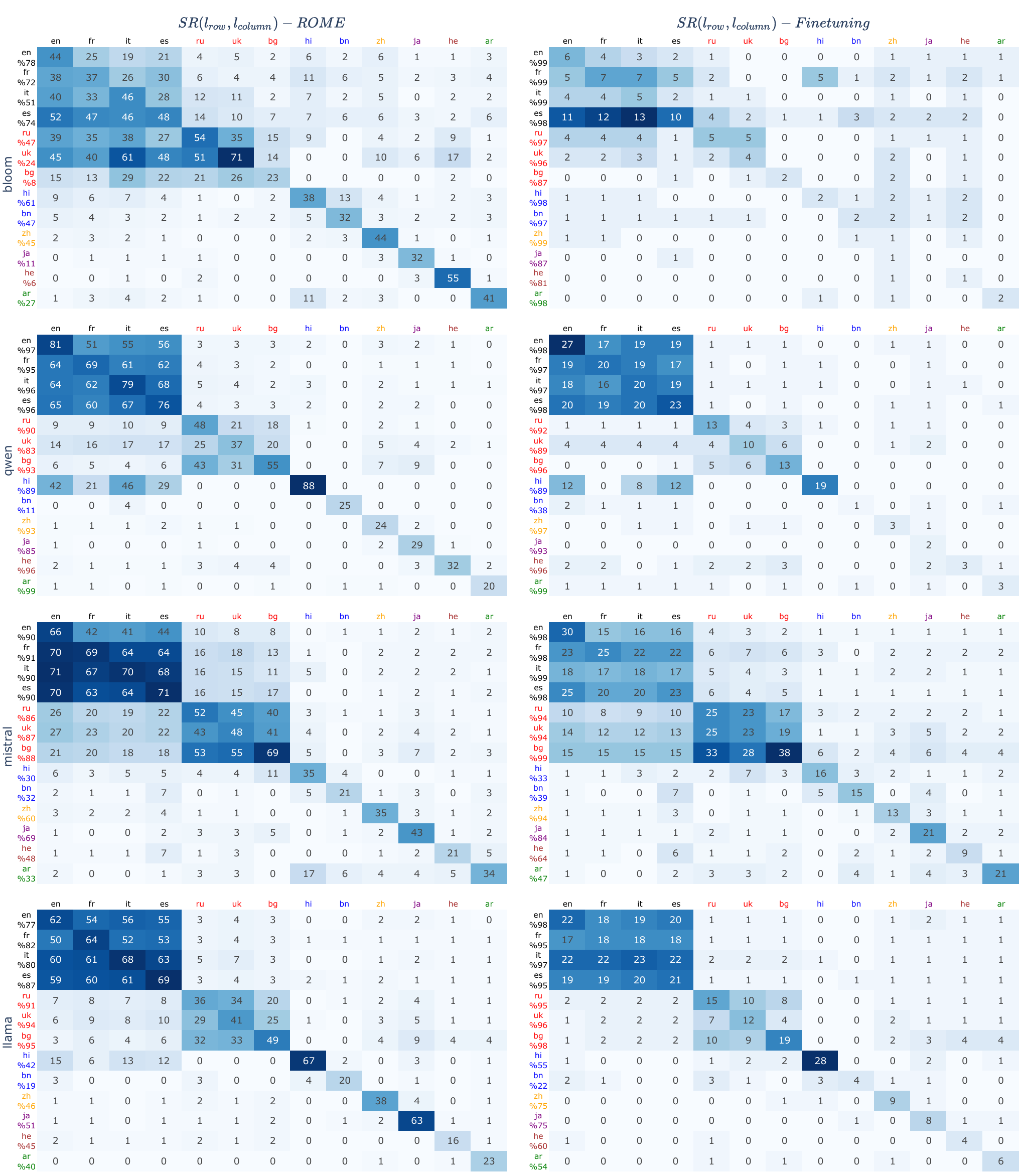}
\end{figure}

\clearpage
\clearpage
\section{On CKR Features}
\label{sec:ckr_features}

\begin{figure}[h]
\centering
\includegraphics[scale=0.5]{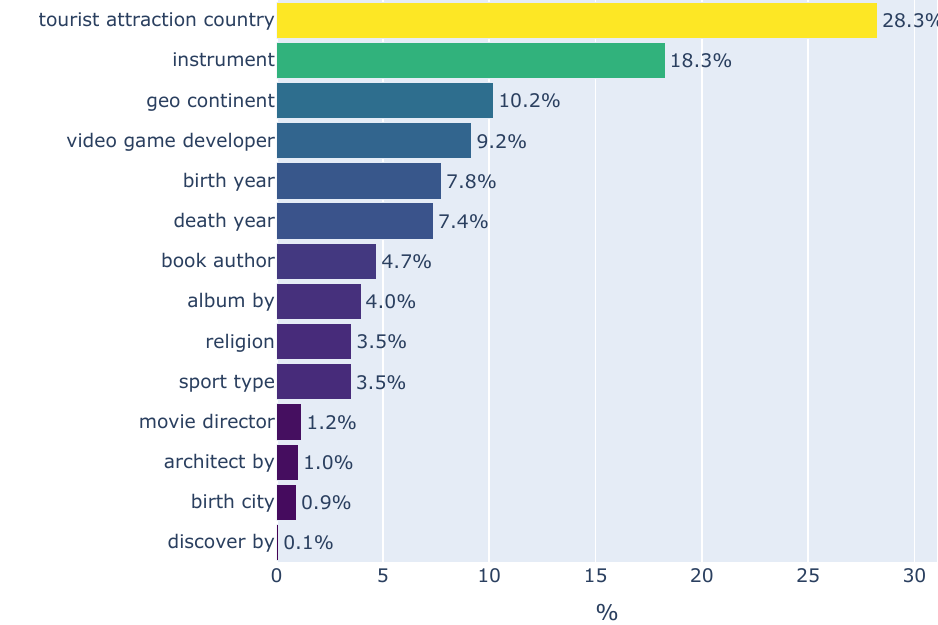}
\caption{Distribution of shared representation facts across different relation types.}
\label{fig:relation_transfer}
\end{figure}

Figure \ref{fig:relation_transfer} illustrates that relations with fewer possible categories generally exhibit higher CKR. This trend is evident for relations such as countries, instruments, continents, and company developers, with sports type and religion as notable exceptions. Numerical relations like birth year and death year also demonstrate strong transfer capabilities. In contrast, relations involving names (e.g., book authors, movie directors, discoverers) and those with numerous categories (e.g., cities) show lower transfer rates.

\end{document}